\documentclass[11pt]{article}

\usepackage[final]{acl}

\usepackage{times}
\usepackage{latexsym}

\usepackage[T1]{fontenc}
\usepackage[utf8]{inputenc}

\usepackage{microtype}

\usepackage{inconsolata}

\usepackage{graphicx}

\usepackage[utf8]{inputenc} % allow utf-8 input
\usepackage{booktabs}       % professional-quality tables
\usepackage{amsfonts}       % blackboard math symbols
\usepackage{nicefrac}       % compact symbols for 1/2, etc.
\usepackage{microtype}      % microtypography
\usepackage{array}   
\usepackage{multirow}
\usepackage{lipsum}            % 用来生成示例文字，可删
\usepackage{amsmath} 
\usepackage[most]{tcolorbox}
\title{Agentic Episodic Control}

\author{
 \textbf{Xidong Yang\textsuperscript{1}},
 \textbf{Wenhao Li\textsuperscript{2}},
 \textbf{Junjie Sheng\textsuperscript{1}},
 \textbf{Yun Hua\textsuperscript{3}},
 \textbf{Haosheng Chen\textsuperscript{1}},\\
 \textbf{Chuyun Shen\textsuperscript{4}\thanks{Corresponding authors.}},
 \textbf{Xiangfeng Wang\textsuperscript{1,5}\footnotemark[1]}
\\
 \textsuperscript{1}East China Normal University\ \ \textsuperscript{2}Tongji University\ \ \textsuperscript{3}Shanghai Jiao Tong University
 \\
 \textsuperscript{4}Shanghai University of International Business and Economics\ \  \textsuperscript{5}Shenzhen Loop Area Institute\\
 \texttt{xdyang@stu.ecnu.edu.cn, cyshen@suibe.edu.cn, xfwang@cs.ecnu.edu.cn}
}

\begin{document}
\maketitle
\begin{abstract}
Reinforcement learning (RL) remains fundamentally limited by poor data efficiency and weak generalization. Prior episodic RL methods attempt to alleviate this via external memory modules, yet they suffer from two key limitations: a representation bottleneck caused by shallow encoders, and a retrieval dilemma where episodic memory is accessed indiscriminately.
To address these challenges, we propose \textbf{Agentic Episodic Control} (AEC), a novel architecture that integrates large language models (LLMs) into episodic RL.
AEC uses an LLM-based semantic augmenter to generate semantic representations from raw observations, and a critical state recognizer to selectively retrieve valuable experiences.
This transforms memory usage from passive similarity matching into strategic, context-aware recall.
Across five BabyAI-Text environments, AEC achieves 2–6$\times$ higher data efficiency than baselines and is the only method to solve complex tasks like UnlockLocal with over 90\% success.
It further demonstrates strong cross-task and cross-environment generalization, maintaining performance even under distribution shifts.
AEC shows that combining LLM-derived priors with reinforcement learning yields more sample-efficient and adaptable agents. Code is available at \url{https://github.com/Xidong-Yang/Agentic_Episodic_Control}.
\end{abstract}

\section{Introduction}

Reinforcement Learning (RL) has been a driving force in AI over the past decade~\cite{silver2016mastering, tunyasuvunakool2021highly, fawzi2022discovering, mankowitz2023faster}.  
Today, RL remains pivotal through Reinforcement Learning from Human Feedback (RLHF) and reinforcement fine-tuning (RFT), which aligns large language models with human preferences~\cite{ouyang2022training, trung2024reft}.
Despite these successes, fundamental challenges separate RL agents from human learners in two critical aspects: \textbf{data efficiency} and \textbf{generalization}~\cite{schwarzer2021pretraining, korkmaz2024survey}.
While humans learn robustly from a handful of experiences and seamlessly transfer knowledge, RL agents often require millions of interactions to master a single task and fail to adapt when faced with novel compositions.
We argue that this gap stems from a deeper deficiency: \textit{the absence of cognitive mechanisms that are central to human learning, particularly a memory system capable of semantic abstraction and strategic retrieval of relevant experiences}~\cite{EICHENBAUM2004109, kuhl2009strategic}.

Recognizing the importance of memory, prior works such as Neural Episodic Control (NEC)~\cite{pritzel2017neural} and episodic reinforcement learning frameworks~\cite{liang2024episodic, na2024efficient} attempted to bridge this gap by incorporating external memory modules.
These methods typically use a trainable neural network to generate state embeddings, which are stored in an episodic memory buffer.
At each decision step, the agent performs a K-nearest-neighbor (KNN) lookup over this buffer to retrieve relevant past experiences—either using them directly to estimate Q-values for action selection, or to provide targets for updating the Q-network.
However, because these designs rely on lightweight state encoders, they yield narrow, task-specific embeddings with insufficient semantic coverage for systematic generalization.
Moreover, in sparse-reward settings, the episodic memory is dominated by near-zero-return trajectories, and per-timestep passive KNN reads therefore tend to retrieve low-value neighbors, which provide neither actionable signals nor useful training targets for the Q-network.
Consequently, these pioneering efforts suffer from two critical flaws: a representation bottleneck, wherein shallow encoders produce retrieval keys with limited semantic and relational abstraction; and a retrieval dilemma, wherein the agent lacks the capacity to decide when or whether to retrieve from memory, defaulting instead to unconditional KNN queries at each step.

Building on the above analysis, we argue that an effective framework still lacks an oracle—a module that, given the current state, infers latent task-relevant factors, completes the state representation accordingly, and decides whether to exploit memory or explore.
Fortunately, we now stand at a unique inflection point, where the rise of large language models (LLMs) provides a practical path to the missing \emph{oracle}.
Trained on massive corpora that implicitly encode broad human knowledge, LLMs act as powerful semantic engines that support high-level reasoning—logical inference~\cite{patel2024multi}, commonsense reasoning~\cite{wang2023gemini}, and abstract problem solving~\cite{imani2023mathprompter}.
Crucially, by conditioning prompts on the current state and task context, we can elicit oracle-like signals: latent, task-relevant abstractions, candidate hypotheses, and decision heuristics that indicate whether to exploit prior experience or explore under distributional shift.
In this sense, LLMs offer a readily accessible proxy for the oracle envisioned above.

This capability directly addresses the representation bottleneck by prompting LLMs with the current state and task context, allowing them to use their own commonsense knowledge to infer additional information beneath the surface state.
Furthermore, their inherent capacity for logical inference and commonsense reasoning provides the foundation for building strategic, context-aware retrieval mechanisms, indicating whether to exploit prior experience or explore under a distributional shift, thus offering a clear path to resolving the retrieval dilemma.

Driven by the ability of LLMs, we introduce Agentic Episodic Control (AEC), a novel framework that addresses the representation bottleneck and retrieval dilemma by integrating two LLM-guided modules.
An LLM-guided augmenter maps raw observations to language-grounded abstractions that expose latent, task-relevant structure.
A critical-state recognizer then determines when consulting episodic memory is warranted and when to act using the agent’s current understanding without recall.
By turning memory access from passive similarity matching into selective, decision-aware recall, AEC yields improved data efficiency and generalization across varied task settings.
Our contributions are four-fold:

\begin{itemize}
    \item We identify two fundamental limitations in episodic RL: a representation bottleneck caused by shallow encoders, and a retrieval dilemma stemming from indiscriminate memory access.
    \item  We propose Agentic Episodic Control (AEC), which addresses both limitations by leveraging large language models to generate abstract state representations and selectively retrieve memory based on task-critical context.
    \item Through experiments on five BabyAI-Text tasks, AEC achieves 2- to 6-fold improvements in data efficiency over baselines, and is the only method to solve complex tasks like UnlockLocal with near-perfect success.
    \item We further validate AEC’s generalization across tasks and environments, showing robust performance under distribution shift and effective memory reuse across settings.
\end{itemize}

\section{Related Work}
\label{related_work}

\subsection{Neural Episodic Control}

Episodic control refers to capturing and retaining the most influential experiences during the learning process to guide an agent's decision-making.
Model-free episodic control (MFEC)~\cite{blundell2016model} employs episodic memory buffers that hold experience tuples, compressing states via Gaussian random projection or variational auto-encoders.
NEC~\cite{pritzel2017neural} introduced a differentiable neural dictionary that stores past experiences, with convolutional networks used for state embedding.
Subsequent studies modify parameterised models through episodic memory~\cite{lin2018episodic, hansen2018fast, chen2022deep, liang2024episodic, lee2019sample}.
Recently, research has shifted to state representation in neural episodic control.
Episodic Reinforcement Learning with Associative Memory (ERLAM)~\cite{zhu2020episodic} associates related trajectories to form policies with efficient reasoning and propagates new values through a graph built over memory items.
Neural Episodic Control with State Abstraction (NECSA)~\cite{li2023neural} performs state abstraction on grid-based environments to exploit topological structures and enhance policy learning and generalization.
% Efficient Episodic Memory Utilization (EMU)~\cite{na2024efficient} constructs semantically coherent memory embeddings with a trainable encoder–decoder.
% Cooperative Embodied Language Agent (CoELA)~\cite{zhangbuilding} maintains three memory systems—semantic, episodic, and procedural—that are updated primarily by logging raw action and dialogue histories.
However, human episodic memory encodes experiences with prior knowledge and semantic detail~\cite{xue2018neural, ramey2022episodic}.
We therefore introduce an LLM-based semantic augmenter, aligned with this human characteristic to enhance decision-making.

\subsection{LLMs for Reinforcement Learning}

Effectively enhancing reinforcement learning with large language models is a promising research area.
Several approaches have been explored to leverage LLMs in this field.
One method is based on a hierarchical framework, in which LLMs are used to decompose complex tasks and generate high-level plans, which are then executed by low-level controllers~\cite{peng2024hypothesis, shukla2023lgts}.
% For example, Hypothesis, Verification, and Induction framework (HYVIN)~\cite{peng2024hypothesis} first employs a large language model to propose subgoal hypotheses.
% Once these hypotheses are validated, general skills are learned under the guidance of these validated subgoals.
Another approach utilizes LLMs to design reward functions, simplifying the typically labor-intensive process of reward function formulation~\cite{kwon2023reward, wu2023read, li2024auto, bhambri2024efficient}.
% For instance, read and reward framework~\cite{wu2023read} significantly improves the learning efficiency and performance of RL agents in Atari games by reading human-written game manuals, extracting key information, and generating auxiliary rewards.
% Auto MC-Reward~\cite{li2024auto} leverages large language models to automatically design dense reward functions, where the reward function is an executable Python function that uses predefined observation inputs to construct the reward.
Some studies directly use LLMs as behavioral policies, training them via RL to interact with the environment~\cite{yan2023ask, shi2023unleashing, li2022pre, carta2023grounding, tan2024twosome}.
For example, Grounded Language Models (GALM)~\cite{carta2023grounding} and Twosome~\cite{tan2024twosome} apply LLMs to text-based games.
Our work differs from these approaches by leveraging the semantic understanding of LLMs to enhance episodic memory embedding, while dynamically coordinating with graph-structured working memory for real-time reasoning.
This tight integration enables more human-like generalization and sample efficiency compared to existing LLM-RL hybrids.

\subsection{LLM-based State Embedding}

Recent advances demonstrate the versatile capabilities of large language models in enriching state representations across diverse applications~\cite{da2024prompt, shek2024lancar, chen2023llm, wang2024llm, benara2024crafting}.
Language for context-aware robot locomotion (LANCAR)~\cite{shek2024lancar} employs an LLM-based contextual information translation module to interpret high-level information from the environment into contextual embeddings accessible to RL agents.
LLM-State~\cite{chen2023llm} leveraged the inherent contextual understanding and historical action reasoning capabilities of LLMs to provide continuous expansion and updates of object properties.
Question-answering embeddings (QA‑Emb)~\cite{benara2024crafting} is a method that leverages LLMs to answer a series of yes/no questions for generating interpretable state embeddings.
Our method integrates a LLM-based semantic encoder directly into the agent"s core memory architecture, enabling deep semantic grounding of both episodic memory storage and retrieval processes.

\section{The AEC Framework}

\label{Framework}
\begin{figure*}[t]
  \centering
  \includegraphics[width=0.8\linewidth]{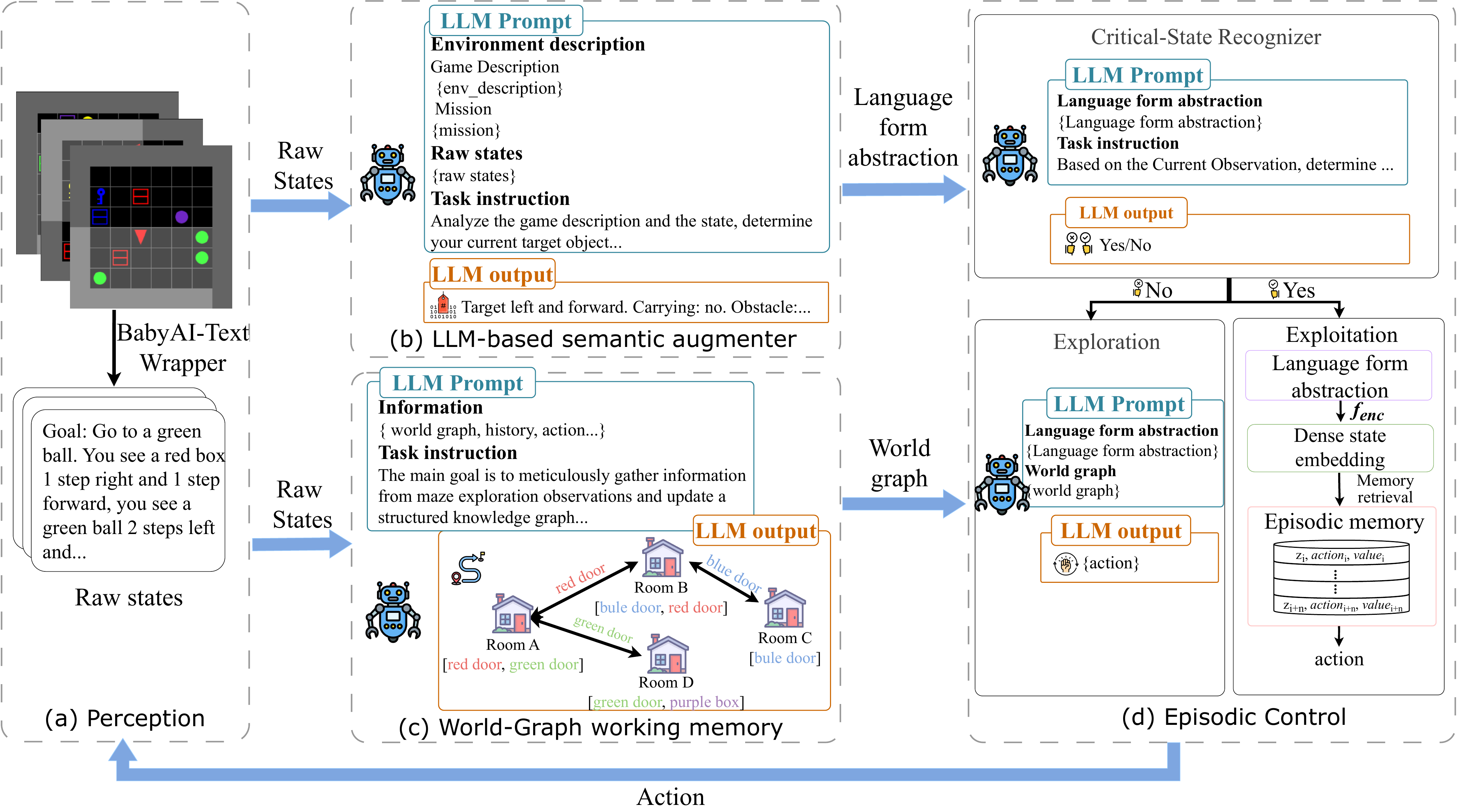}
  \caption{The overview of Agentic Episodic Control framework. Raw states are processed in parallel by an LLM-based semantic augmenter (b) to produce language-form abstraction and by a World-Graph working memory module (c) to build a structured working memory of entities and relations. An episodic control module (d), equipped with a critical-state recognizer, then uses these representations to decide whether to exploit episodic memory or to continue exploring under World-Graph guidance. The chosen actions interact with the environment.}
  \label{fig:overview}
\end{figure*}

We propose an efficient episodic control framework called \textbf{Agentic Episodic Control} (see Figure~\ref{fig:overview}), designed to simultaneously address the representation bottleneck and the retrieval dilemma of NEC and its variants.
To overcome the representation bottleneck, AEC incorporates a semantic augmenter based on LLMs, which maps raw observations into language-aligned abstract representations.
This yields richer feature embeddings and more stable keys for memory operations.
To tackle the retrieval dilemma, AEC introduces a working memory that maintains a short-term summary of the current context to support exploration, along with a critical-state recognizer that dynamically mediates between exploitation and exploration.
Episodic recall is selectively triggered only when it is likely to significantly influence future decisions; otherwise, the agent proceeds based on its current internal model.
This transforms memory access from passive similarity-based retrieval into targeted, decision-aware recall, thereby enhancing data efficiency and generalization.

\noindent \textbf{Preliminary.}
Episodic control draws on the hippocampus’s memory mechanism~\cite{EICHENBAUM2004109}. 
At each timestep \(t\), the agent observes a state \(s_t\).
To enable memory-based decision-making, the agent employs a state embedding function \(f_{\phi}(s): \mathcal{S} \rightarrow \mathbb{R}^k\), which maps raw states into a \(k\)-dimensional latent space~\cite{na2024efficient}.
The resulting embedding is denoted by \(z_t = f_{\phi}(s_t)\).

For each action, the agent maintains an episodic memory buffer \(\mathcal{D}_E^a\), which stores key–value pairs where each key is a state embedding \(z_t\), and each value stores the highest return observed for a given state–action pair.
Formally, given a pair \((s_t, a_t)\), the agent updates the value \(H(z_t)\) within the specific buffer \(\mathcal{D}_E^{a_t}\) according to~\cite{lin2018episodic}:
% \begin{equation}
% \begin{aligned}
% H(z_t) =
% \begin{cases}
% \max\!\left\{ H(\hat{z}_t),\, R_t(s_t, a_t) \right\}, \\
% \hspace{2em} \text{if } (s_t, a_t) \in \mathcal{D}_E^a, \\[4pt]
% R(s_t, a_t), \\
% \hspace{2em} \text{otherwise}.
% \end{cases}
% \end{aligned}
% \label{eq:episodic_update}
% \end{equation}
%省空间
\begin{equation}
\small
H(z_t) =
\begin{cases}
\max\!\left\{ H(\hat{z}_t),\, R_t(s_t, a_t) \right\}, & \text{if } (s_t, a_t) \in \mathcal{D}_E^{a_t}, \\[4pt]
R(s_t, a_t), & \text{otherwise},
\end{cases}
\label{eq:episodic_update}
\end{equation}

where \(R(s_t, a_t)\) denotes the return received after taking action \(a_t\) in state \(s_t\), and \(\hat{z}_t\) represents the exact match of the current state–action pair in memory, if available.

At decision time, the agent queries the episodic memory for each candidate action \(a\). Specifically, it performs a $k$-nearest neighbor (KNN) search over \(\mathcal{D}_E^a\) using the current state embedding \(z_t\), yielding a set of top-$k$ closest entries \(\mathcal{N}_p(z_t, \mathcal{D}_E^a)\). Each retrieved value is weighted according to its similarity to the query embedding, producing a kernel-weighted action value estimate:

\begin{equation}
Q(s_t, a) = \sum_{i \in \mathcal{N}_p(z_t, \mathcal{D}_E^a)} w_i \cdot H(z_i),
\end{equation}

where the weights \(w_i\) are typically computed via a softmax over cosine similarities or a kernel function.

Finally, the selected action is:
\begin{equation}
a_t = \arg\max_{a} Q(s_t, a).
\end{equation}

This memory-based approach enables rapid value estimation from past experience and underpins the enhanced episodic control mechanisms introduced in this work.

\subsection{How to Solve the Representation Bottleneck?}
\label{LLM_based_state_encoder}

To address the representation bottleneck, we introduce an LLM-based semantic augmenter that enriches raw inputs by generating language-based abstractions that capture deeper contextual meaning.
These abstractions expose latent task-relevant features, enabling robust memory indexing.
In this section, we describe how these representations are generated, stored, and queried to support more efficient and generalizable episodic memory.

\noindent \textbf{Representation Generation via LLMs.}
We construct semantic state representations by integrating commonsense knowledge through an LLM-based semantic augmenter as $LLM_{\text{aug}}$.
Specifically, given a raw state $s$, the augmenter takes a structured prompt that includes the environment description, the raw observation, and the task instruction.
It outputs a language-form abstraction.
\begin{equation}
\mathbf{l}(s)=LLM_{\text{aug}}(s),
\end{equation}
This module is implemented via prompting and does not require additional fine-tuning. 
A complete template is given in the Appendix~\ref{app:prompt_engineering}.
To enable efficient retrieval, we further compute a dense state embedding $\mathbf{z}(s_t)$ via a fixed pretrained text encoder $f_{\text{enc}}$~\cite{reimers-2019-sentence-bert}
\begin{equation}
\mathbf{z}(s_t)=f_{\text{enc}}(\mathbf{l}(s)),
\end{equation}
which encapsulates the textual representation of the state.
These representations are more semantically meaningful and provide more information, making them ideal for indexing and retrieval in episodic memory.

% \noindent \textbf{Episodic Memory Construction and Update.}
\noindent \textbf{Representation Storage and Memory Update.}
To implement episodic memory control, we maintain an episodic memory \(\mathcal{M}\) that stores representations in triplets, each triplet consisting of a state embedding, the action that has historically achieved the highest observed return, and its return:
\begin{equation}
\mathcal{M} = \big(\mathbf{z}(s),\, a^{*},\, \hat{Q}(s, a^{*}) \big),
\end{equation}
\begin{equation}
a^{*} = \arg\max_{a} \hat{Q}(s,a),
\end{equation}
where \(a^{*}\) is the action with the highest recorded return from \(s\), and \(\hat{Q}(s,a^{*})\) is the maximum value.

At each timestep $t$, the agent observes the current state $s_t$ and stores the corresponding state embedding $\mathbf{z}(s_t)$, the taken action $a$, and the received reward $r$ in an episodic data buffer.
Only after the episode ends is $\hat{Q}$ computed based on the collected data, and the episodic memory is then updated accordingly.
We insert new entries for unseen states, and update existing ones only if the current return exceeds the stored value.
% \begin{equation}
% \mathcal{M}_{t+1}(\mathbf{z}(s))=
% \begin{cases}
% \big(\mathbf{z}(s),\, a,\, \hat{Q}(s,a)\big), & \mathbf{z}(s)\notin \mathcal{M}_t,\\[6pt]
% \big(\mathbf{z}(s),\, a^{*},\, \hat{Q}(s, a^{*})\big), & \text{otherwise},
% \end{cases}
% \end{equation}
% where
% $$
% (a^{*}, \hat{Q}(s, a^{*}))=\underset{(a,v)\in\{(a,\hat{Q}(s,a),\,(a^{\text{EM}}_t(s),\,Q^{\text{EM}}_t(s))\}}{\max}\; v.
% $$

\noindent \textbf{Memory Retrieval.}
We perform a KNN search in the dense state embedding space with $k = 1$. The LLM-based semantic augmenter enriches the representation with contextual meaning and latent information beyond the surface-level observation, enabling the embedding space to cluster \emph{functionally similar} states. As a result, retrieving a single nearest neighbor is often sufficient and helps avoid irrelevant or misleading matches. This clustering effect arises from expressing goal-relevant relations and affordances in natural language, which are then encoded into dense vector representations.
Let \(\mathbf{z}_t \) be the query embedding at time $t$ and
\(\mathbf{z}_i\) be the embedding stored with the \(i\)-th memory entry
\((\mathbf{z}_i, a_i^{*}, \hat{Q}(s_i,a_i^{*})) \in \mathcal{M}\).
We measure similarity with cosine similarity:
\begin{equation}
\mathrm{sim}(\mathbf{z}_t, \mathbf{z}_i) = \frac{\mathbf{z}_t^{\top}\mathbf{z}_i}{\|\mathbf{z}_t\|_2 \, \|\mathbf{z}_i\|_2}.
\end{equation}
The \(k\)-nearest neighbors of \(s\) are then
\begin{equation}
\mathcal{N}_k(s) = \operatorname{TopK}_{i \in \mathcal{M}} \; \mathrm{sim}(\mathbf{z}_t, \mathbf{z}_i),
\end{equation}
and in our experiments we set \(k = 1\).
The retrieved memory tuple is \((\mathbf{z}_i, a_{i}^{*}, \hat{Q}(s_{i}, a_{i}^{*}))\).

\subsection{How to Address the Retrieval Dilemma?}

To resolve the retrieval dilemma, we introduce a two-part mechanism: (i) a World-Graph working memory that maintains structured summaries for short-term reasoning, and (ii) a critical-state recognizer that selectively triggers memory recall at decision-critical moments. Together, these modules transform retrieval from passive matching into active, context-aware decisions.

\noindent \textbf{World-Graph Working Memory.}
Prior work has shown that combining working memory with episodic memory can improve generalization in reinforcement learning agents~\cite{fortunato2019generalization}.
Building on this insight, we introduce a World-Graph working memory to maintain a structured representation of the agent’s interaction history.
The environment is represented as a dynamic graph
\begin{equation}
\mathcal{G} = (\mathcal{V}, \mathcal{E}, \mathcal{X}),
\end{equation}
where each node $v \in \mathcal{V}$ corresponds to a distinct and persistent environment configuration.
Edges $e \in \mathcal{E}$ represent actions that induce transitions between configurations, and $\mathcal{X}$ stores configuration-specific attributes.
This design allows the graph to compactly encode the agent’s accumulated knowledge of environment dynamics.

% At each timestep $t$, the World-Graph module updates $\mathcal{G}_t$ based on the current state $s_t$ and past interactions.
% If the current state corresponds to a previously unseen environment configuration, a new node is added; otherwise, the existing node is reused.
% Edges are created or updated to reflect action-induced transitions, while node attributes are refined as new information becomes available.
At each timestep $t$, the World-Graph module updates $\mathcal{G}_t$ based on the current state $s_t$ and past interactions, adding a new node only when a previously unseen environment configuration is encountered.
Edges are created or updated to reflect action-induced transitions, while node attributes are refined as new information becomes available.
For example, in navigation tasks, a new node is created only when the agent enters a previously unexplored room, whereas movements within the same room update the current node.
Through this incremental construction process, the World-Graph captures the agent’s evolving understanding of environment dynamics and provides structured support for reasoning and decision-making.

\noindent \textbf{Critical-State Recognizer.}
\begin{figure*}[bt]
\centering
\includegraphics[width=1.0\linewidth]{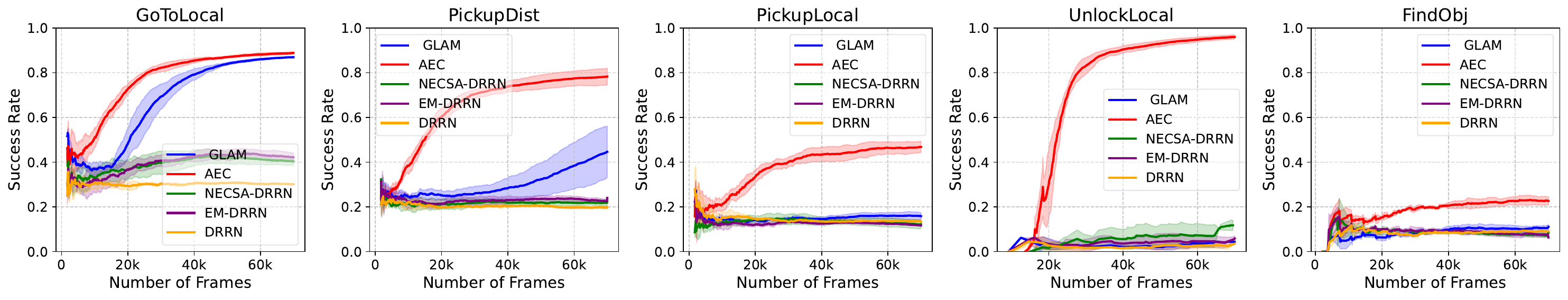} 
\caption{
Learning curves (success rate vs. training frames) for five BabyAI-Text tasks comparing AEC with four baseline methods.
For each method the solid line shows the mean success rate over three seeds, and the surrounding shaded region marks one standard deviation above and below that mean.
AEC consistently demonstrates faster convergence and higher final performance across tasks.}
\label{fig:mainresult.}
\end{figure*}
While the World-Graph working memory offers a real-time spatial abstraction of the current environment, and the episodic memory preserves valuable past experiences, the retrieval dilemma remains unresolved: when should the agent consult memory, and when should it rely on its current understanding? To address this, we introduce a critical-state recognizer, powered by a large language model, which identifies moments when episodic recall is likely to influence future outcomes. By detecting such decision-critical states, the agent can selectively trigger memory retrieval only when it is beneficial, thereby improving efficiency and avoiding unnecessary or distracting recall.
This mechanism ensures that historical experience is injected precisely at important decision points, transforms memory access from passive similarity-based retrieval into targeted, decision-aware recall.

\noindent\textbf{Definition 1.}
(Critical State) A state \(s_t\) is critical if the choice of action at \(s_t\) can substantially change the remaining cumulative return.
Formally,
\begin{equation}
    \Delta(s_t)=\max_{a}{Q}(s_t,a)-\min_{a}{Q}(s_t,a).
\end{equation}
A state is typically classified as critical when $\Delta(s_t) > \tau_Q$, where $\tau_Q$ is a preset hyperparameter.

Computing \(\Delta(s_t)\) requires a trained Q-network and a tuned threshold \(\tau_Q\), which is brittle and task-specific. Instead, we use a large language model to directly decide criticality from the semantic abstraction \(LLM_{\text{aug}}(s_t)\), enabling generalization without retraining or hyperparameter tuning:
\begin{equation}
    c(s_t)=LLM_{\text{crit}}(LLM_{\text{aug}}(s_t))\in\{0,1\}.
\end{equation}
\noindent \textbf{Exploitation.}
If the current state is classified as critical, the agent triggers an episodic memory retrieval.
The agent queries the episodic memory with $f_{\text{enc}}(LLM_{\text{aug}}(s_t))$.
As described in Section~\ref{LLM_based_state_encoder}, we perform an exact KNN search ($k = 1$) over $\mathcal{M}$.
If a matching entry exists, the agent executes the stored action; otherwise, it samples an action at random:
\begin{equation}
    a_t =
\begin{cases}
a_{i}^{*}, & \text{if } (\mathbf{z}_{i}, a_{i}^{*}, \hat{Q}(s_{i}, a_{i}^{*})) \in \mathcal{M},\\[4pt]
\text{Uniform}(\mathcal{A}), & \text{otherwise},
\end{cases}
\end{equation}
where \(\mathcal{A}\) denotes the action space.

\noindent \textbf{Exploration.}
If the state is not critical, the agent proceeds with exploration.
Given the current state \(s_t\) and the up‑to‑date World-Graph working memory \(\mathcal{G}_t\), it chooses the next action by reasoning over these two sources of information.
In practice, this step is implemented by a decision LLM that takes \((s_t, \mathcal{G}_t)\) as input and outputs an action.

\section{Experiments}

\begin{table*}[t]
\centering
\resizebox{\linewidth}{!}{
\begin{tabular}{llccccc}
\toprule
\textbf{Setting} & \textbf{Method} & \textbf{GoToLocal} & \textbf{PickupDist} & \textbf{PickupLocal} & \textbf{UnlockLocal} & \textbf{FindObj} \\
\midrule
\multirow{5}{*}{No Change} 
    & DRRN & 0.13 ± 0.02 & 0.14 ± 0.02 & 0.01 ± 0.02 & / &  0.03 ± 0.06 \\
    & EM-DRRN & 0.24 ± 0.09 & 0.15 ± 0.02 & 0.04 ± 0.01 & / &  0.03 ± 0.06 \\
    & NECSA-DRRN & 0.18 ± 0.02 & 0.05 ± 0.03 &  0.01 ± 0.01 & / &  / \\
    & GLAM & \textbf{0.91 ± 0.08} & 0.69 ± 0.03 &  0.18 ± 0.04 & 0.01 ± 0.01 &  0.13 ± 0.06 \\
    & AEC & 0.84 ± 0.02  & \textbf{0.75 ± 0.17} & \textbf{0.45 ± 0.02} & \textbf{0.95 ± 0.04} &  \textbf{0.23 ± 0.02} \\
\midrule
\multirow{5}{*}{New Object} 
    & DRRN & 0.08 ± 0.02{$\downarrow$0.05} & 0.11 ± 0.03{$\downarrow$0.03} &  /{$\downarrow$0.01}  & / {$\bullet$}  &  /{$\downarrow$0.03} \\
    & EM-DRRN & 0.10 ± 0.07{$\downarrow$0.14} & 0.09 ± 0.09{$\downarrow$0.06} &  /{$\downarrow$0.04}  & / {$\bullet$} &  /{$\downarrow$0.03} \\
    & NECSA-DRRN & 0.19 ± 0.03{$\uparrow$0.01} & 0.02 ± 0.02{$\downarrow$0.03} &  0.02 ± 0.00{$\uparrow$0.01} & / {$\bullet$} &  / {$\bullet$} \\
    & GLAM & \textbf{0.85 ± 0.10}{$\downarrow$0.06} & 0.61 ± 0.04{$\downarrow$0.08} & 0.16 ± 0.06{$\downarrow$0.02} & 0.01 ± 0.01{$\bullet$} &  0.12 ± 0.03{$\downarrow$0.01} \\
    & AEC & 0.83 ± 0.02{$\downarrow$0.01} & \textbf{0.71 ± 0.12}{$\downarrow$0.04} &  \textbf{0.52 ± 0.06}{$\uparrow$0.07} & \textbf{0.95 ± 0.02}{$\bullet$} &  \textbf{0.20 ± 0.03}{$\downarrow$0.03} \\
\bottomrule
\end{tabular}
}
\caption{
Final success rates (mean ± std over 3 seeds) after 70K environment frames on five BabyAI-Text environments, evaluated under two settings: \textbf{No Change} (seen objects) and \textbf{New Object} (unseen objects). ``/'' indicates the method failed to solve the task. Performance differences between the two settings are indicated as follows: ${\uparrow}$ (increase), ${\downarrow}$ (decrease), and ${\bullet}$ (no change).}
\label{tab:performance_comparison}
\end{table*}

We assess AEC through the following research questions: 

\noindent \textbf{RQ1 (Data Efficiency)}: Can AEC improve data efficiency?

\noindent \textbf{RQ2 (Generalization)}: How well does AEC generalize across different tasks and environments?

\noindent \textbf{RQ3 (Representation Bottleneck)}: Does LLM-based semantic augmenter produce better state representations for episodic memory retrieval?

\noindent \textbf{RQ4 (Retrieval Dilemma)}: Is selective episodic retrieval more effective than indiscriminate memory access?

\subsection{Experimental Setup}

\noindent \textbf{Environment.}
% We evaluate the proposed Agentic Episodic Control architecture on the BabyAI-Text benchmark~\cite{carta2023grounding}, a textual variant of the BabyAI environment~\cite{chevalier2018babyai} that reformulates grid-based instruction-following tasks into a purely text-based format.
% We adopt BabyAI-Text as our evaluation platform because it provides a controlled yet challenging testbed for studying grounded language understanding, semantic decision-making, and generalization in RL agents.
We evaluate Agentic Episodic Control on the BabyAI-Text benchmark~\cite{carta2023grounding}, a textual variant of BabyAI~\cite{chevalier2018babyai} that reformulates grid-based instruction-following tasks into a controlled yet challenging testbed for grounded language understanding, semantic decision-making, and generalization in reinforcement learning.

Our evaluation covers five BabyAI-Text environments that probe complementary capabilities: \textit{GoToLocal}, \textit{PickupDist}, \textit{PickupLocal}, \textit{UnlockLocal}, and \textit{FindObj}.% ACL 添加appendix的引用
Following the protocol in~\cite{carta2023grounding}, we also test the agent’s ability to generalize to unseen instructions and object types.
We refer to the standard setting as \textit{No change} and the version containing novel, unseen target objects as \textit{New object}.  %添加appendix的引用
More details are given in Appendix~\ref{app:env_detail}.
AEC is built on top of the \texttt{Qwen2.5-32B-Instruct}~\cite{qwen2.5} model, a powerful large language model with strong instruction-following capabilities.

% \paragraph{Baselines}
% We validate the effectiveness of AEC by comparing it with several baselines spanning episodic memory reinforcement learning and fine-tuned large language model methods. {\bf DRRN}~\cite{he2016deep} serves as a strong RL baseline with near‑state‑of‑the‑art performance on many medium‑ to hard‑level interactive text environments~\cite{wang2022scienceworld}. {\bf EM-DRRN} is an episodic memory-augmented DRRN. Our implementation follows a similar approach to EMDQN~\cite{lin2018episodic}. {\bf NECSA-DRRN} is also an episodic memory-augmented method. NECSA~\cite{li2023neural} works by discretizing the state-action space into a grid structure, enabling efficient policy learning; {\bf GLAM}~\cite{carta2023grounding} grounds a large language model via online PPO fine-tuning, allowing it to act directly as the policy in an interactive text environment. Our AEC is built upon the Qwen2.5-32B-Instruct~\cite{qwen2.5} large language model, leveraging its advanced language understanding and generation capabilities.

\noindent \textbf{Baselines.}
We compare AEC against a range of baselines, covering both reinforcement learning and large language model–augmented approaches.
\textbf{DRRN}~\cite{he2016deep} serves as a standard RL baseline with competitive performance in interactive text environments~\cite{wang2022scienceworld}, including the BabyAI-Text benchmark~\cite{carta2023grounding}.
\textbf{EM-DRRN} extends DRRN with an episodic memory module, following the design of EMDQN~\cite{lin2018episodic}, to improve sample efficiency.
\textbf{NECSA-DRRN} incorporates NECSA~\cite{li2023neural} into DRRN, discretizing the state-action space into a grid-like structure for efficient memory-based value estimation.
\textbf{GLAM}~\cite{carta2023grounding} fine-tunes a large language model with PPO to act as a policy directly in textual environments, serving as a strong LLM-based baseline.
% Our method, \textbf{AEC}, is built upon the \texttt{Qwen2.5-32B-Instruct}~\cite{qwen2.5} model, which provides powerful language understanding and generation capabilities. AEC leverages this foundation through semantic embedding, episodic retrieval, and world-model–guided reasoning to enhance decision-making in partially observable, language-driven tasks.

\subsection{RQ1: Data Efficiency}

We benchmarked AEC against four baselines on five BabyAI-Text environments.
As shown in Figure~\ref{fig:mainresult.}, AEC consistently achieves higher success rates with significantly fewer interactions.
In the obstacle navigation environment GoToLocal, AEC surpasses an 80\% success rate within 25K frames, while GLAM requires approximately 50K frames to reach a comparable level (75\%), demonstrating nearly a 2× improvement in sample efficiency.
Although EM-DRRN and NECSA-DRRN benefit from episodic memory to slightly accelerate learning, they still fall far behind AEC.
In the object pickup environment PickupDist, for example, GLAM requires 60K frames to reach a 40\% success rate, whereas AEC achieves this performance within just 10K frames—representing a 6× gain in data efficiency.
For the multi-step reasoning environment UnlockLocal, AEC is the only method capable of solving the task, surpassing a 90\% success rate within 40K frames. All baselines fail to exceed 15\% even after 70K frames, highlighting AEC’s strong generalization and planning capabilities.
In the most challenging environment, FindObj, which involves multi-room exploration under sparse rewards, AEC again stands out as the only method to exceed a 20\% success rate within 70K frames.

Taken together, these results clearly demonstrate that AEC significantly improves data efficiency across a diverse set of language-conditioned reasoning and navigation tasks.

\subsection{RQ2: Generalization}

\noindent \textbf{Cross-Task.}
We evaluate AEC’s generalization ability across different tasks settings by comparing its performance under two evaluation conditions: No Change, where the agent encounters familiar objects seen during training, and New Object, where novel target objects are introduced at test time.

As shown in Table~\ref{tab:performance_comparison}, AEC demonstrates strong generalization across all five BabyAI-Text tasks. While it performs slightly below GLAM on GoToLocal, it consistently outperforms all baselines on the remaining tasks under both evaluation settings. Notably, AEC is the only method that maintains high performance even when evaluated with previously unseen objects. Despite the distribution shift introduced in the New Object setting, AEC retains most of its performance. For example, on PickupLocal, AEC even exceeds its performance under the No Change condition, achieving a relative improvement of +16\% (from 0.45 to 0.52). In more challenging scenarios like FindObj, AEC still retains 87\% of its original performance (dropping only slightly from 0.23 to 0.20), whereas other baselines degrade significantly or fail entirely.
These results confirm that AEC’s integration of language-grounded abstraction and structured memory enables it to generalize effectively to novel task configurations with minimal performance loss.

\noindent \textbf{Cross-Environment.} We investigate whether AEC’s episodic memory enables cross-environment generalization.
To this end, we conducted a transfer experiment in which the agent attempted to solve a task using episodic memories retrieved from a different environment.
Specifically, we evaluated performance on the GotoLocal and PickupLocal environments under three memory settings: raw states from the same environment, memory from the same environment, and memory from a different environment.

As shown in Table~\ref{tab:cross_task_memory}, reusing memory from a related environment leads to notable improvements over using raw states.
For example, using PickupLocal-derived memory to solve GotoLocal yields a success rate of 0.75, substantially higher than the 0.49 obtained with raw states.
Similarly, applying GotoLocal memory to PickupLocal improves performance from 0.30 to 0.39.
Interestingly, we observe an asymmetry in transferability: experiences learned from the more complex PickupLocal environment benefit GotoLocal more than the reverse.
This suggests that solving more demanding environments may yield richer, more generalizable memory representations, providing stronger priors when applied to simpler environments.
These results indicate the potential for cross-environment memory reuse in AEC and open avenues for future work on compositional generalization.

\begin{table}[t]
\small
\resizebox{\linewidth}{!}{
\centering
\begin{tabular}{lcc}
\toprule
\multirow{2}{*}{\textbf{Memory Source}} & \multicolumn{2}{c}{\textbf{Target Task}} \\
                               & \textbf{GotoLocal} & \textbf{PickupLocal} \\
\midrule
\textbf{Raw States}         & 0.49 & 0.30 \\
\textbf{GotoLocal Memory}   & 0.83 & 0.39 \\
\textbf{PickupLocal Memory} & 0.75 & 0.52 \\
\bottomrule
\end{tabular}
}
\caption{Cross-task memory transfer results. Success rates on GotoLocal and PickupLocal tasks when using episodic memory encoded from different tasks. ``Raw States'' denotes raw states without LLM-based semantic augmenter from the same task.}
% Results show that memory from related tasks improves performance.}
\label{tab:cross_task_memory}
\end{table}

\subsection{RQ3: Representation Bottleneck}

\begin{figure}[t]
\centering
\includegraphics[width=0.9\columnwidth]{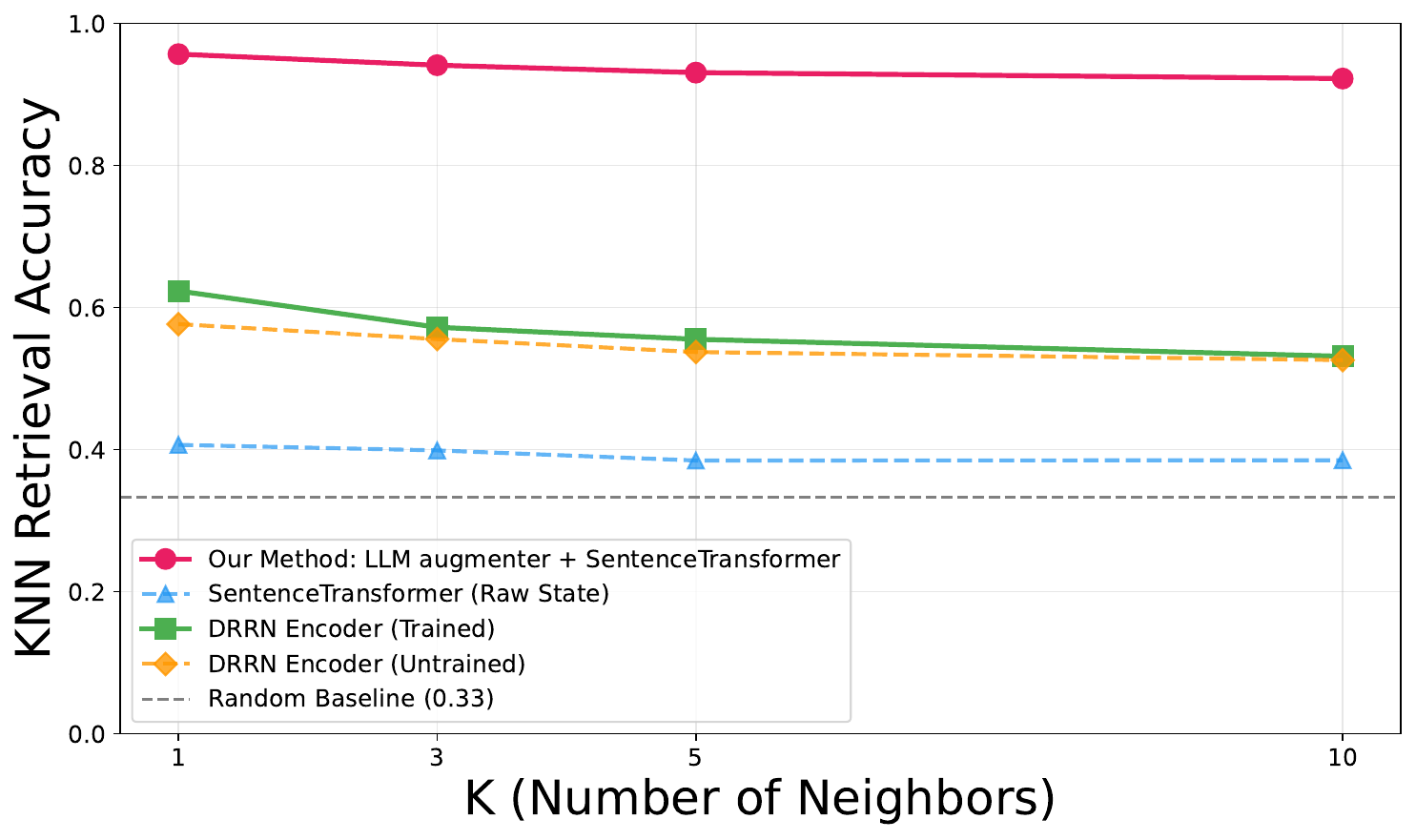} % Reduce the figure size so that it is slightly narrower than the column. Don't use precise values for figure width.This setup will avoid overfull boxes.
\caption{KNN retrieval accuracy for different state representations.}
\label{fig:knn_comparison}
\end{figure}

To evaluate whether our method alleviates the representation bottleneck in episodic reinforcement learning, we perform a controlled representation analysis on states with known decision semantics.
We construct a balanced set of 300 states by randomly sampling 100 states for each of three manually identified optimal actions, enabling a direct test of whether similarity in the embedding space aligns with \emph{decision equivalence}.

We compare four state representations: SentenceTransformer embeddings of LLM-augmented semantic representations, SentenceTransformer embeddings of raw textual states, embeddings from a DRRN encoder trained for 70K environment frames, and embeddings from an untrained DRRN encoder.
For each representation, we perform KNN retrieval and measure \emph{KNN retrieval accuracy}, defined as the proportion of retrieved neighbors that share the same optimal action.

As shown in Figure~\ref{fig:knn_comparison}, LLM-based representations achieve consistently high retrieval accuracy across all values of $k$, exceeding 95\% at $k=1$ and remaining above 92\% at $k=10$.
DRRN representations trained for 70K frames show only a modest improvement over untrained DRRN embeddings.
We note that with substantially more training data and interaction, DRRN-style encoders may learn stronger task-specific representations.
However, under the low-data regime considered here, shallow and partially trained encoders fail to align embedding similarity with decision relevance, whereas LLM-based semantic augmenter yields highly action-consistent neighborhoods.
% These results demonstrate that our method effectively alleviates the representation bottleneck in episodic reinforcement learning.

\subsection{RQ4: Retrieval Dilemma}
% To investigate the impact of the retrieval dilemma on decision quality, we compare AEC against a variant that queries episodic memory indiscriminately at every timestep (AEC w/o selective retrieval) across three environments: GoToLocal, PickupLocal, and UnlockLocal.
% Crucially, both methods utilize the identical episodic memory content to ensure a fair comparison.
% Figure~\ref{fig:retrieval_dilemma} reports both task success rates and the memory retrieval ratio, defined as the proportion of timesteps in which episodic memory is accessed.

\begin{figure}[t]
\centering
\includegraphics[width=0.9\columnwidth]{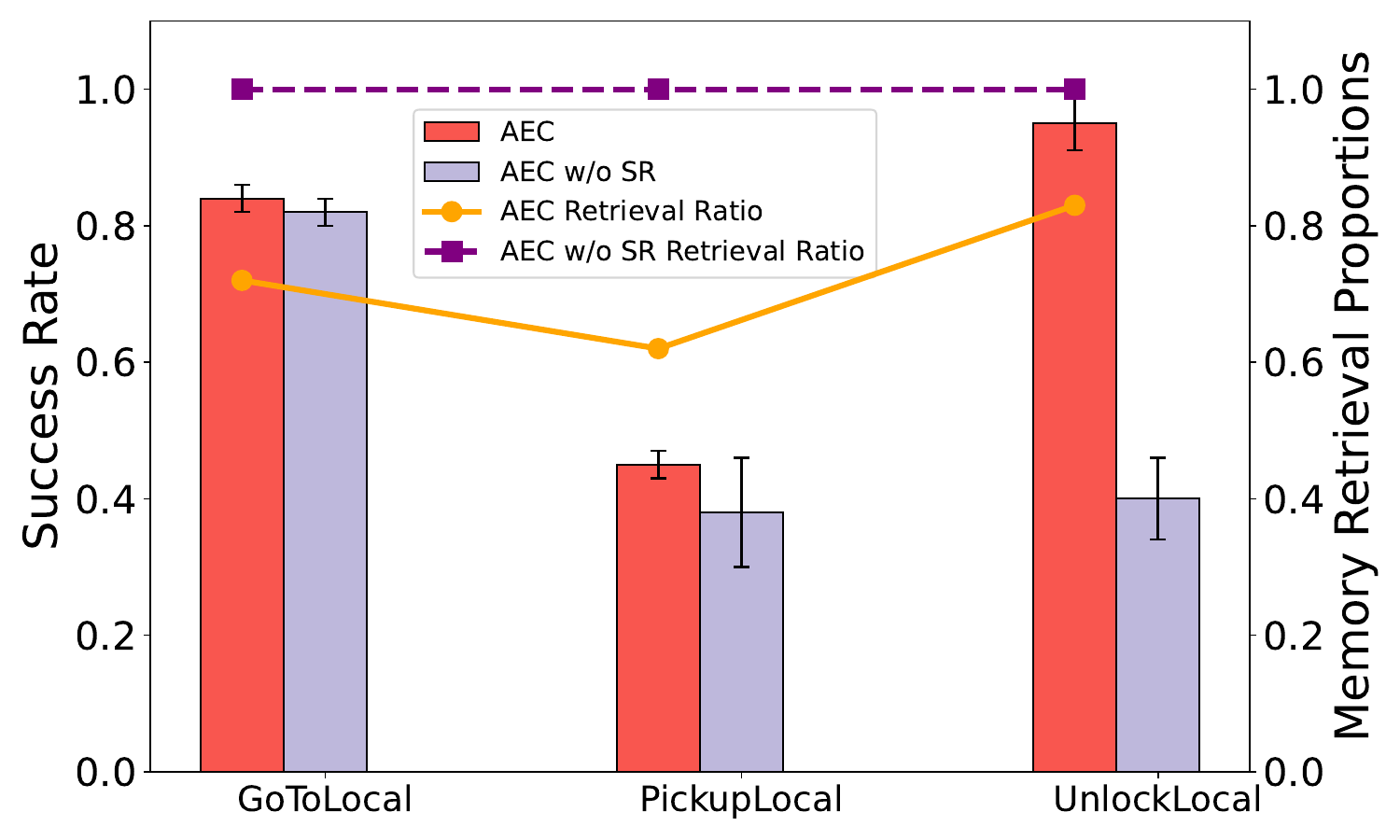} % Reduce the figure size so that it is slightly narrower than the column. Don't use precise values for figure width.This setup will avoid overfull boxes.
\caption{Comparison of success rates and memory retrieval proportions on three BabyAI-Text environments. SR denotes Selective Retrieval. Error bars represent one standard deviation.}
\label{fig:retrieval_dilemma}
\end{figure}

% Across all environments, AEC achieves higher success rates while significantly reducing memory queries.
% The improvement is most evident in UnlockLocal, where only a small number of states are decision-critical.
% These results indicate that indiscriminate retrieval often introduces irrelevant experiences, whereas selectively invoking episodic memory at critical states leads to more effective decision-making.
% This demonstrates that resolving the retrieval dilemma improves both performance and efficiency in episodic reinforcement learning.
To investigate the impact of the retrieval dilemma on decision quality, we compare AEC against a variant that queries episodic memory indiscriminately at every timestep (AEC w/o selective retrieval).
Crucially, both agents utilize the identical episodic memory content to ensure a fair comparison.
Figure~\ref{fig:retrieval_dilemma} reports task success rates and memory retrieval proportions across three environments: GoToLocal, PickupLocal, and UnlockLocal.

AEC consistently outperforms the indiscriminate baseline across all tasks.
This performance gap is particularly pronounced in UnlockLocal, where the baseline's success rate drops significantly.
These results indicate that accessing episodic memory at every step is detrimental to decision accuracy, potentially due to the interference caused by retrieving information at non-critical states.
Consequently, selectively invoking episodic memory is essential for maximizing task success and robustness in complex environments.

\begin{table}[t]
\resizebox{\linewidth}{!}{
\centering
\begin{tabular}{lccc}
\toprule
\textbf{LLM} & \textbf{GoToLocal} & \textbf{PickupLocal} & \textbf{FindObj}\\
\midrule
\textbf{Qwen2.5-7B-Instruct}      & 0.79 ± 0.03 & 0.43 ± 0.02 &0.13 ± 0.01  \\
\textbf{Qwen2.5-32B-Instruct}      & 0.84 ± 0.02 & 0.45 ± 0.02 & 0.23 ± 0.02\\
\bottomrule
\end{tabular}
}
\caption{Robustness across different LLM backbones on the three BabyAI-Text tasks (100 test environments). 
Values are means over three seeds; $\pm$ denotes one standard deviation.}
\label{tab:llm_robustness}
\end{table}

\begin{figure}[t]
\centering
\includegraphics[width=0.9\columnwidth]{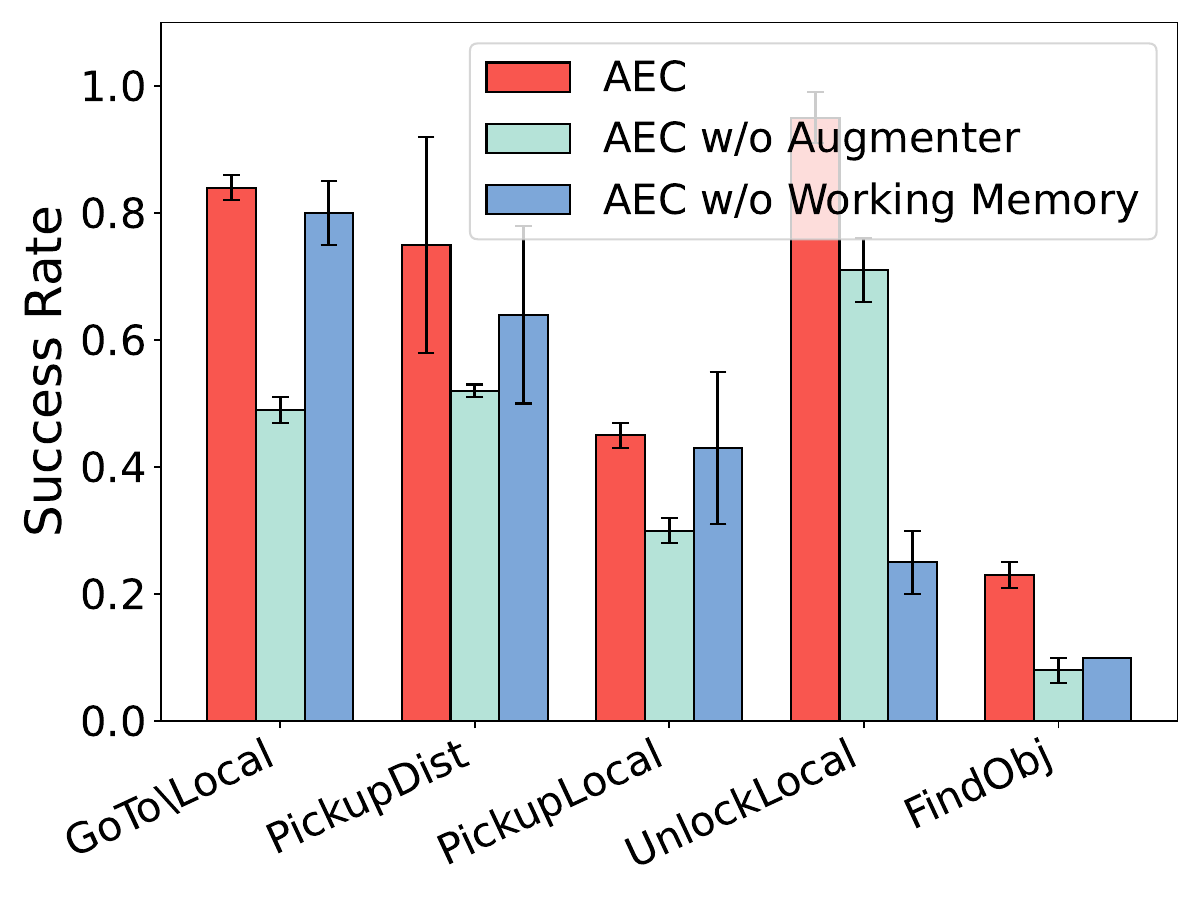} % Reduce the figure size so that it is slightly narrower than the column. Don't use precise values for figure width.This setup will avoid overfull boxes.
\caption{Ablation study results across five BabyAI-Text environments. Error bars represent one standard deviation.}
\label{fig:ablation}
\end{figure}

\subsection{Ablation studies}

% We conducted an ablation study across all five BabyAI-Text environments, focusing on the effects of the component in AEC. As shown in Figure~\ref{fig:ablation}, removing either module leads to a consistent performance drop across all tasks, highlighting their complementary roles. For example, working memory proves especially critical in FindObj, a task characterized by extremely sparse rewards, where the agent receives little useful feedback from the environment. This highlights the importance of effectively resolving the retrieval dilemma under such conditions—ensuring that memory is consulted only when it can meaningfully guide decision-making. Together, these results confirm that both the LLM-based semantic augmenter and the working memory are essential for AEC’s strong performance, enabling it to generalize semantically while reasoning over extended horizons.

% To assess AEC’s robustness across backbone models, we reran the experiments with the smaller \texttt{Qwen2.5-7B-Instruct}~\cite{qwen2.5} model on three tasks of clearly different difficulty, using three random seeds per task. Each score reported in Table~\ref{tab:llm_robustness} is the mean over 100 test environments. The results show that AEC performs consistently well with both LLMs, confirming its robustness and stability. Nevertheless, the agent powered by \texttt{Qwen2.5-32B-Instruct} still surpasses the 7B model, suggesting that AEC further benefits from the richer knowledge embedded in larger language models.

We conduct ablation studies on all five BabyAI-Text environments to assess the contribution of AEC’s components.
As shown in Figure~\ref{fig:ablation}, removing either module consistently degrades performance across tasks, demonstrating their complementary roles.
In particular, working memory is crucial for FindObj, which features extremely sparse rewards, where selective memory retrieval is essential to avoid uninformative recall.
These results confirm that both the LLM-based semantic augmenter and the working memory are necessary for AEC’s strong performance, enabling semantic generalization and long-horizon reasoning.

To evaluate robustness across backbone models, we repeat experiments using the smaller \texttt{Qwen2.5-7B-Instruct} model on three tasks of varying difficulty, with three random seeds per task.
Table~\ref{tab:llm_robustness} reports mean success rates over 100 test environments.
AEC achieves consistent performance with both backbones, while the 32B model performs better overall, indicating that AEC further benefits from the richer knowledge of larger language models.

\section{Conclusions}

We introduced Agentic Episodic Control, a framework that integrates LLM-based semantic abstraction with structured memory to resolve the representation bottleneck and retrieval dilemma in episodic RL. By replacing passive retrieval with decision-aware recall, AEC achieves significant improvements in sample efficiency and generalization, consistently outperforming strong baselines in complex, long-horizon tasks. These results underscore the effectiveness of combining semantic understanding with active memory mechanisms, offering a robust path toward more data-efficient autonomous agents.

\section*{Limitations}

Despite its strong performance, AEC incurs higher inference latency compared to traditional RL baselines due to the computational cost of LLM queries and structured memory updates. This overhead may limit its deployment in real-time, high-frequency control settings. Future work will focus on mitigating these costs through techniques such as knowledge distillation into smaller models or caching intermediate semantic representations.

\section*{Acknowledgments}

Wenhao Li is supported by the NSFC (62406270) and the STCSM Shanghai Rising-Star Program (24YF2748800).
Yun Hua is supported by the Shanghai Post-doctoral Excellence Program (2025251).
Xiangfeng Wang is supported by the SHEITC (2025-GZL-RGZN-BTBX-01004).

% This document has been adapted
% by Steven Bethard, Ryan Cotterell and Rui Yan
% from the instructions for earlier ACL and NAACL proceedings, including those for
% ACL 2019 by Douwe Kiela and Ivan Vuli\'{c},
% NAACL 2019 by Stephanie Lukin and Alla Roskovskaya,
% ACL 2018 by Shay Cohen, Kevin Gimpel, and Wei Lu,
% NAACL 2018 by Margaret Mitchell and Stephanie Lukin,
% Bib\TeX{} suggestions for (NA)ACL 2017/2018 from Jason Eisner,
% ACL 2017 by Dan Gildea and Min-Yen Kan,
% NAACL 2017 by Margaret Mitchell,
% ACL 2012 by Maggie Li and Michael White,
% ACL 2010 by Jing-Shin Chang and Philipp Koehn,
% ACL 2008 by Johanna D. Moore, Simone Teufel, James Allan, and Sadaoki Furui,
% ACL 2005 by Hwee Tou Ng and Kemal Oflazer,
% ACL 2002 by Eugene Charniak and Dekang Lin,
% and earlier ACL and EACL formats written by several people, including
% John Chen, Henry S. Thompson and Donald Walker.
% Additional elements were taken from the formatting instructions of the \emph{International Joint Conference on Artificial Intelligence} and the \emph{Conference on Computer Vision and Pattern Recognition}.


\bibliography{latex/aaai2026}
\clearpage
\appendix

\section{Ethical Considerations}
This study investigates the role of large language models as components within reinforcement learning agents, and all experimental results are obtained by evaluating agent performance in controlled environments.
In addition, an LLM was used as an auxiliary tool during manuscript preparation, including improving grammar and clarity of the text.
The LLM was not used to generate experimental data, design algorithms, analyze results, or draw scientific conclusions.
All research ideas, methods, experiments, analyses, and conclusions were conceived and conducted by the authors.

\section{Environment Details}
\label{app:env_detail}

We evaluate the proposed Agentic Episodic Control architecture on the BabyAI-Text benchmark~\cite{carta2023grounding}.
The wrapper~\cite{carta2023grounding} built on top of BabyAI removes reliance on visual perception and allows us to isolate the benefits of language-driven episodic memory and structured reasoning.
In this environment, the agent is provided with a natural language instruction and a textual description of its local egocentric observation. This observation represents a forward-facing view in text form, covering up to seven adjacent tiles directly in front of the agent.
This design simulates partial observability and tests the agent’s ability to reason over limited, language-grounded sensory input.

Our evaluation covers five \textbf{BabyAI-Text} environments that probe complementary capabilities:
\begin{itemize}
    \item \textit{GoToLocal}: The agent must navigate to a specified object in a room with distractors, assessing spatial grounding and distractor resistance.

    \item \textit{PickupDist}: Requires picking up a specified object (by type and/or color) amid distractors, stressing fine-grained object disambiguation.

    \item \textit{PickupLocal}: Builds on \textit{PickupDist}, but describes the target via its location, testing grounding of locative language in a single-room setting.

    \item \textit{UnlockLocal}: Involves fetching the correct key and unlocking a door in the current room, emphasizing multi-step reasoning and tool use.

    \item \textit{FindObj}: Evaluates exploration, requiring the agent to explore across six rooms to locate a target object.
\end{itemize}

Following the protocol in~\cite{carta2023grounding}, we also test the agent’s ability to generalize to unseen instructions and object types.  
We refer to the standard evaluation setting as \textit{No change}, where the object names and colors present in the environment during testing are consistent with those seen during training.
However, the target instruction and room layout vary across episodes.
In contrast, the \textit{New object} setting introduces novel, previously unseen object names and colors at test time, while still varying the target and room layout in each episode.

\section{Case Studies}

\begin{figure}[tb!]
\centering
\includegraphics[width=1.0\linewidth]{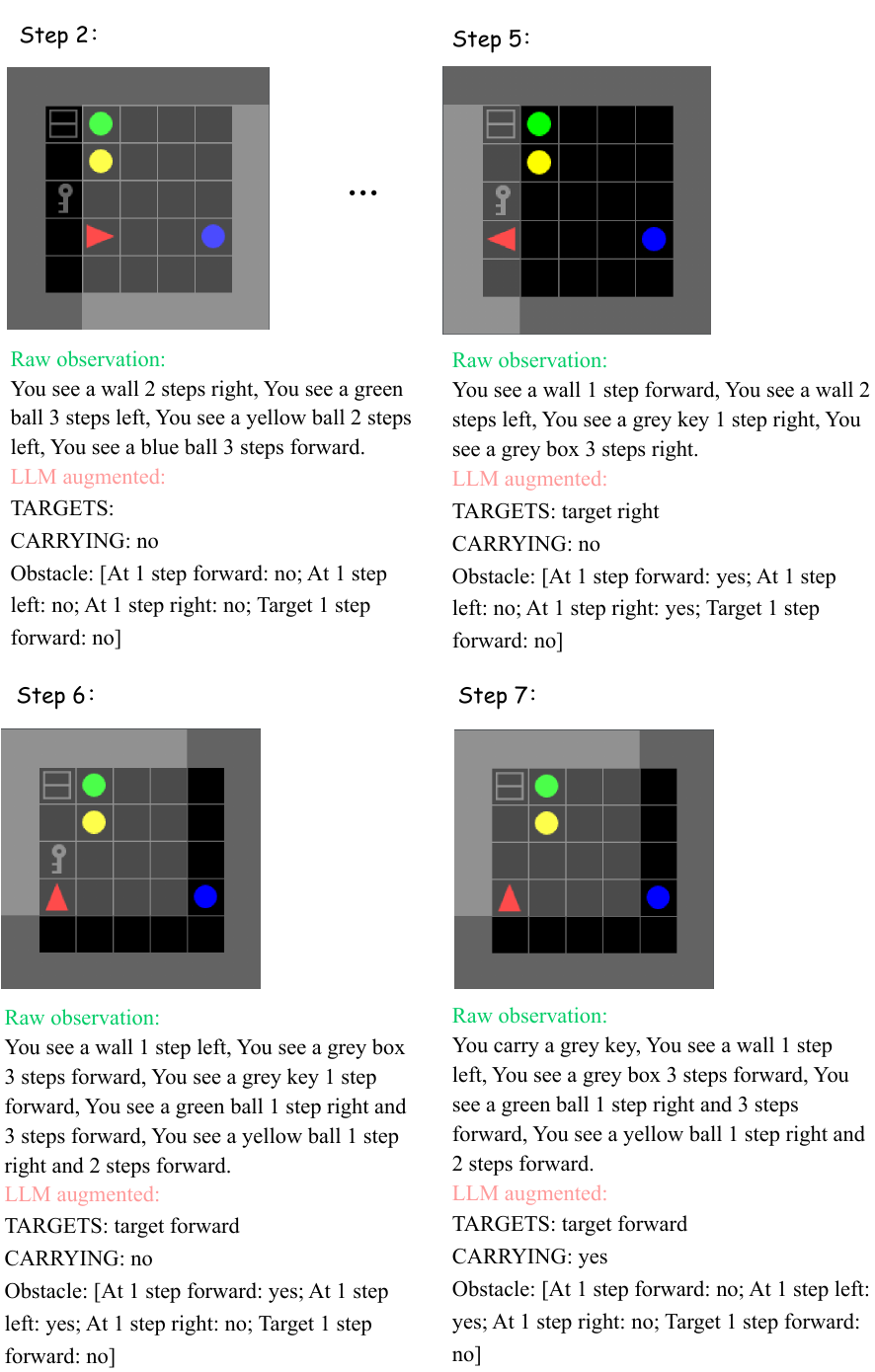} 
\caption{
Case study of the LLM-based semantic augmenter in the PickupDist task (\emph{pick up the gray box}).
For selected timesteps, we show the textual description, and the LLM-augmented semantic representations.}
\label{fig:state_aug}
\end{figure}

\begin{figure}[tb!]
\centering
\includegraphics[width=1.0\linewidth]{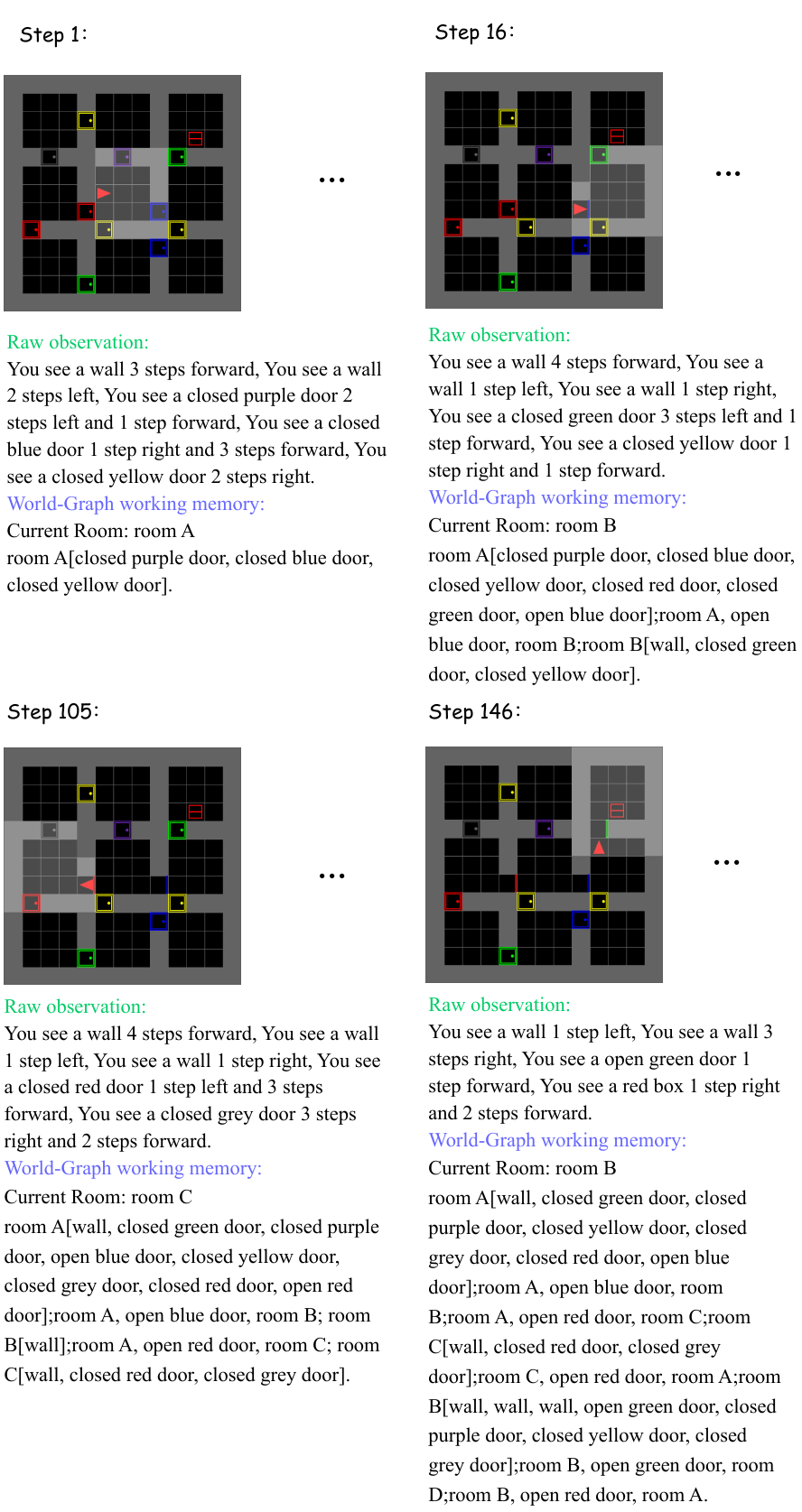} 
\caption{
Selected key steps from the “pick up the box” task. Each frame illustrates the agent’s textual description, and the corresponding state of its World-Graph working memory. Since the environment does not provide explicit room names, the agent assigns room labels (e.g., room A, room B) based on its own exploration order and infers connectivity between rooms using door colors and relative positions.}
\label{fig:wg}
\end{figure}

\begin{figure}[tb!]
\centering
\includegraphics[width=1.0\linewidth]{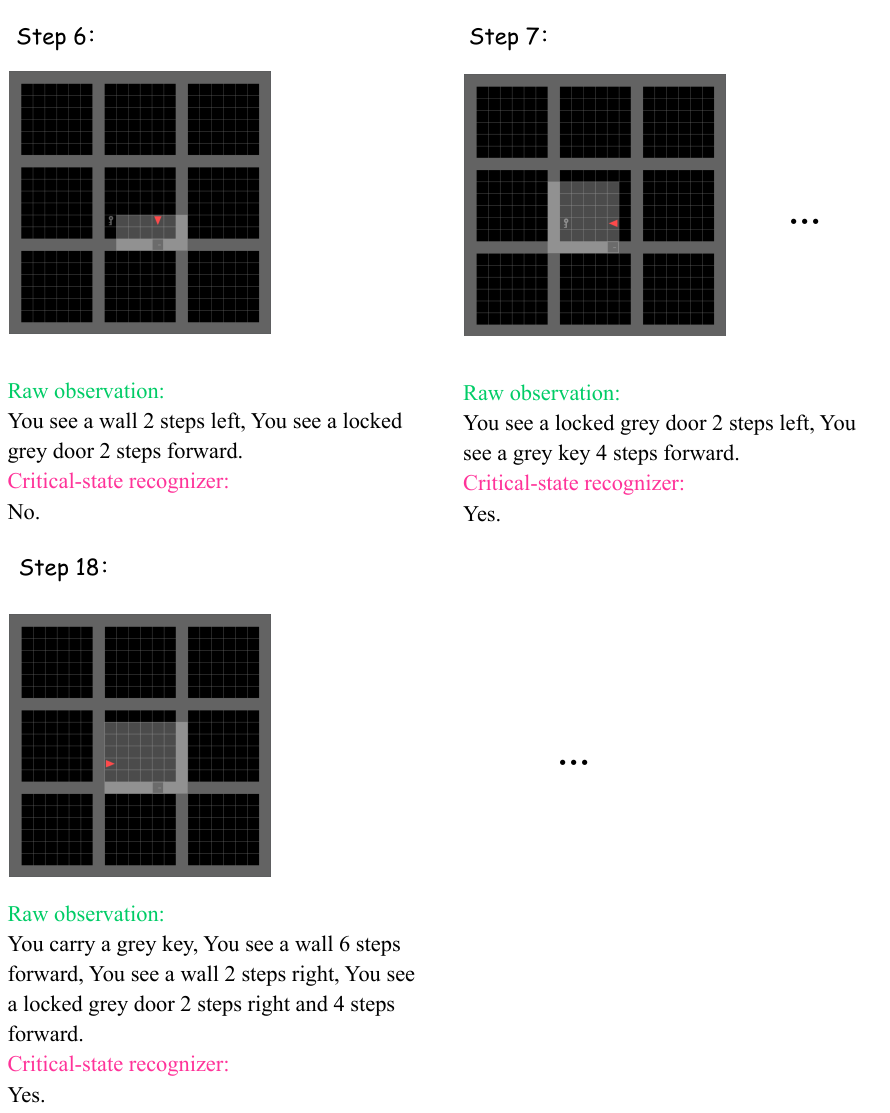} 
\caption{
Performance of the critical-state recognizer in the UnlockLocal environment.}
\label{fig:cs}
\end{figure}

\noindent \textbf{Analysis of LLM-based semantic augmenter.}
To illustrate the role of the LLM-based semantic augmenter, we analyze a representative episode from the PickupDist task, focusing on how raw observations are transformed into structured, decision-relevant representations.

As shown in  Figure~\ref{fig:state_aug}, raw observations consist of unstructured, surface-level descriptions that enumerate visible objects and their relative positions. While informative, these observations entangle task-relevant signals with irrelevant details, making it difficult for similarity-based retrieval to identify functionally equivalent states.

In contrast, the LLM-based semantic augmenter distills each observation into a compact representation that explicitly highlights task-critical factors, including the relative direction of the target , the agent’s carrying status, and local obstacle availability.
For example, at Step~5, the augmenter correctly identifies the target as being to the right and marks the immediate right position as blocked by an obstacle, while suppressing unrelated objects such as balls that do not affect decision-making.
As the episode progresses (Steps~6–7), the semantic representation remains stable despite changes in raw visual content, while dynamically updating key attributes such as the agent’s carrying state after picking up the key.

This abstraction process effectively maps perceptually different but functionally similar states to nearby representations, aligning the embedding space with decision relevance rather than surface appearance.
As a result, episodic memory retrieval operates over semantically meaningful keys, enabling reliable reuse of past experience across variations in object layout and observation wording.
This case study demonstrates that the LLM-based semantic augmenter plays a crucial role in alleviating the representation bottleneck by transforming raw observations into task-aligned, decision-centric state representations.

\noindent \textbf{Analysis of World-Graph Working Memory.}
To evaluate the correctness of the World-Graph working memory constructed by the agent during task execution, we analyzed a representative episode from the FindObj task, specifically the “pick up the box” scenario. As illustrated in the Figure~\ref{fig:wg}, we highlight several key steps in the episode, showing the visual observations, textual descriptions, and the corresponding state of the agent’s World-Graph.

Our analysis shows that the agent is capable of accurately building and maintaining connectivity information between rooms during exploration. For instance, by step 146, the World-Graph correctly reflects that Room A is connected to Room B via a blue door, to Room C via a red door, and that Room B is connected to Room D through a green door. This structured spatial representation allows the agent to plan paths and navigate toward the goal room effectively. 

It is important to note that since the environment does not provide explicit room identifiers, the agent assigns room names based on the order of exploration and relies on room attributes to recognize its current location. This approach poses challenges, especially in dynamic environments where, for instance, door states may change—sometimes leading to attribute misidentification and local inconsistencies in the World-Graph. Nevertheless, despite these uncertainties, the agent is still able to infer room connectivity accurately and reliably identify paths to the target, demonstrating robustness in spatial memory construction and planning.

\noindent \textbf{Analysis of Critical-State Recognizer.}
The Figure~\ref{fig:cs}, based on the UnlockLocal environment, illustrates the effectiveness of the critical-state recognizer. In the "open the door" task, before the agent obtains the key, the recognizer identifies observations containing the key as critical states. After acquiring the key, it shifts to recognizing locked doors as critical. This dynamic adaptation shows that the recognizer can adjust its definition of criticality in real time according to the agent’s task progress. As a result, it more effectively guides the agent’s balance between exploration and exploitation, transforming retrieval from passive matching into active, context-aware decision-making—thereby addressing the retrieval dilemma.

\section{Prompt Engineering}
\label{app:prompt_engineering}
We introduce the four prompt templates employed in our experiments.
These prompts are carefully crafted to provide structured and consistent instructions that guide the large language model’s behavior within the text-based maze environment.
Their design aims to enhance clarity, ensure uniform interpretation of observations, support coherent memory management, and promote rational decision-making.
The complete content of each prompt is presented below.
% 定义一个带标题的彩色圆角文本框环境

\noindent\hrulefill

\medskip
\noindent\textit{The prompt for LLM-based semantic augmenter}

\smallskip
\noindent
{\ttfamily\footnotesize
You are MazeNavigator-GPT, an expert at reading text-based maze observations.

GAME DESCRIPTION

\{Game description\}

MISSION
 
\{Mission\}

INPUT

\{Observation\}

TASK (Do all in order, think step‑by‑step)

STEP 0 - PARSE MISSION 

Analyze the game description and the observation, determine your current target object. If the prerequisites for certain goals are not met, they should not be set as current target object. Extract the exact color + item that defines the current target object. Store it as one pair (e.g. "red ball"). This is the current target object.

STEP 1 - BUILD OUTPUT LIST 

Try to find the current target object in the observation. Scan the single observation for any mention of that current target object with the same color. Ignore objects whose does not match exactly. For each valid sentence: remove all numbers (distances, counts, step sizes). Keep the pure direction phrase (e.g. "left and forward", "right"). Create an entry in the form: "target \textless direction phrase \textgreater". Example → "target left and forward". Preserve the order they appear in the observation.

STEP 2 - CARRYING ITEM? 

Output "yes" if the observation states the agent is carrying anything; else "no".

STEP 3 - OBSTACLE ANALYSIS

Is there any objects or walls are there in the three directions of the agent: "1 step forward without left and right", "1 step left without forward", "1 step right without forward"? If no information provided, output "no".

STEP 4 - TARGET ANALYSIS

Target 1 step forward? Output "Target 1 step forward: yes" if the current target object only at 1 step forward; else if the current target object is at 1 step left/right and 1 step forward, output "Target 1 step forward: no".

OUTPUT FORMAT

\{Format requirements\}
}

\noindent\hrulefill

\medskip
\noindent\textit{The prompt for World-Graph working memory construction}

\smallskip
\noindent
{\ttfamily\footnotesize
OBJECTIVE

The main goal is to meticulously gather information from maze exploration observations and update a structured knowledge graph. Don"t include any speculation. Only incorporate new, directly observed information.

Always retain all items and triplets from the previous knowledge graph. Do not remove or modify any previously recorded objects or relationships, even if they are not observed in the current observation.

PREVIOUS WORLD MODEL

\{Previous world model\}

PREVIOUS OBSERVATION

\{Previous observation\}

ACTION

\{Action\}

NEW OBSERVATION

\{New observation\}

TASK

Building and Updating the Knowledge Graph - Follow these steps:

STEP 0 - Room Labeling and Node Properties

Assign room labels sequentially as rooms are confirmed. Name the first room "room A", and label subsequent new rooms as "room B", "room C", etc. If a room has been labeled previously, reuse that label consistently. For the room you have just passed, include all explicitly observed objects in its properties. IMPORTANT: first, import any objects already memorized for that room from the previous world model. Then, add only the newly observed objects (if they are not already recorded). Do not remove any objects from previous observations.- Ignore object positions or orientations. Format: room A [item 1, item 2, item 3, ...]; ...

At the beginning of your output, include a line stating the current room label. Example: "Current Room: room A".

STEP 1 - Triplet Creation

Only perform this step if the new observation explicitly mentions a door. If the new observation does not explicitly mention a door, SKIP THIS STEP entirely and do not create any triplets. When creating triplets, strictly use the format: subject, relation, object. Here the subject is the current room. The relation is the specific door (using the color and state as explicitly mentioned in the observation). The object is the new room reached by that door.

OUTPUT FORMAT

\{Format requirements\}
}

\noindent\hrulefill

\medskip
\noindent\textit{The prompt for critical-state recognizer}

\smallskip
\noindent
{\ttfamily\footnotesize
You are MazeNavigator-GPT, a specialist in solving navigation games.

GAME DESCRIPTION

\{Game description\}

MISSION

\{Mission\}

CURRENT OBSERVATION

\{Current observation after augmentation\}

TASK

Based on the current observation, determine if there is a target direction. If there is, output "yes". If not, output "no". Do not provide any additional explanations, comments, or text.
}

\noindent\hrulefill

\medskip
\noindent\textit{The prompt for exploration strategy using World-Graph working memory}

\smallskip
\noindent
{\ttfamily\footnotesize
You are MazeNavigator-GPT, a specialist in solving navigation games by selecting the optimal action based on the mission, world model, and observations. You enjoy exploring unknown environments.

GAME DESCRIPTION

\{Game description\}

MISSION

\{Mission\}

ACTION SPACE:

\{Action space\}

WORLD MODEL

\{World model\}

HISTORY

\{History\}

CURRENT OBSERVATION

\{Current observation\}

TASK

Based on the world model, history and current observation, select the single most effective action that brings you closer to the mission goal or advances environmental exploration(if no mission objective is present). Be careful to avoid falling into a simple loop of exploration solely through turning left or right;try considering moving forward, picking up items, or other actions to break the stalemate. If you notice that repeating the same actions in consecutive similar scenarios fails to make progress, prioritize trying other options.

OUTPUT FORMAT

\{Format requirements\}

}

\section{Comparison with Advanced Prompt Engineering}

We conducted additional experiments using advanced prompt engineering frameworks, specifically ReAct~\cite{yao2022react} and Tree of Thoughts (ToT)~\cite{yao2023tree}. To ensure a strictly fair comparison, these baselines use the exact same backbone model as our proposed method, \texttt{Qwen2.5-32B-Instruct}, and were evaluated across three environments of varying difficulty.

\begin{table}[h]
\small
\centering
\begin{tabular}{lccc}
\hline
\textbf{Method} & \textbf{GoToLocal} & \textbf{PickupLocal} & \textbf{FindObj} \\
\hline
\textbf{ReAct} & 0.35 & 0.19 & 0.19 \\
\textbf{ToT}   & 0.36 & 0.25 & 0.20 \\
\textbf{AEC (Ours)} & \textbf{0.84} & \textbf{0.45} & \textbf{0.23} \\
\hline
\end{tabular}
\caption{Comparison with advanced prompt engineering baselines.}
\label{tab:e1_prompt_baselines}
\end{table}

These results indicate that even when equipped with advanced prompting techniques, standard LLM agents still struggle with the long-horizon planning and sparse-reward nature of these tasks. In contrast, AEC is specifically designed to address these challenges through its integrated memory architecture.

\end{document}